\documentclass{article}
\usepackage[utf8]{inputenc}
\usepackage{microtype}
\usepackage{graphicx}
\usepackage{subfigure}
\usepackage{booktabs} %
\usepackage{multirow}
\usepackage{hyperref}
\usepackage{chngcntr}
\usepackage{amsmath}
\usepackage{amssymb}

\usepackage[accepted]{icml2020}

\icmltitlerunning{On uncertainty estimation in active learning for image segmentation}

\begin{document}

\twocolumn[
\icmltitle{On uncertainty estimation in active learning for image segmentation}

\begin{icmlauthorlist}
\icmlauthor{Bo Li}{bo,dtu}
\icmlauthor{Tommy Sonne Alstrøm}{dtu}
\end{icmlauthorlist}

\icmlaffiliation{bo}{Department of Information technology, University of Ghent, Belgium}
\icmlaffiliation{dtu}{Department of Applied Mathematics and Computer Science, Technical University of Denmark, Denmark}

\icmlcorrespondingauthor{Bo Li}{Bo.Li@UGent.be}

\icmlkeywords{Machine Learning, ICML}
\vskip 0.3in
]

\printAffiliationsAndNotice{}  %

\begin{abstract}
    
    Uncertainty estimation is important for interpreting the trustworthiness of machine learning models in many applications. This is especially critical in the data-driven active learning setting where the goal is to achieve a certain accuracy with minimum labeling effort. In such settings, the model learns to select the most informative unlabeled samples for annotation based on its estimated uncertainty. The highly uncertain predictions are assumed to be more informative for improving model performance. In this paper, we explore uncertainty calibration within an active learning framework for medical image segmentation, an area where labels often are scarce. Various uncertainty estimation methods and acquisition strategies (regions and full images) are investigated. We observe that selecting regions to annotate instead of full images leads to more well-calibrated models. Additionally, we experimentally show that annotating regions can cut 50\% of pixels that need to be labeled by humans compared to annotating full images. 
\end{abstract}

\section{Introduction}

We address the problem of uncertainty estimation in active learning with application to image segmentation~\cite{Cohn1996}. As the classifier, we use convolution neural networks (CNN) which enables a powerful feature representation capacity~\cite{NIPS2012_4824}. CNNs has gained popularity in histopathological image analysis, where the morphology of histological structures, such as glands and nuclei, needs to be assessed by pathologists to identify the malignancy degree various conditions~\cite{DBLP:journals/tbe/XuLWLFLC17, DBLP:conf/cvpr/0011QYH16, 6944230}. Accurate segmentation and identification is an essential component to obtain reliable morphological statistics for quantitative diagnosis~\cite{DBLP:conf/cvpr/0011QYH16, DBLP:conf/miccai/YangZCZC17}. Recent advances in deep learning using CNNs have achieved promising results in many biomedical image segmentation benchmarks such as nuclei segmentation~\cite{7103332, DBLP:conf/miccai/VetaDP16} and 
gland cell segmentation~\cite{DBLP:conf/cvpr/0011QYH16, DBLP:conf/miccai/XuLLWLC16, 7493349}.

CNNs require annotated data in order to be trained for segmentation, and for medical image segmentation, this incurs a high annotation cost since the annotation must be carried out by specialists. To reduce the labeling cost, there is a pressing need for finding a set with a minimum number of labeled images to achieve a certain segmentation accuracy. Active learning is one of the frameworks that can address this challenge~\cite{Cohn1996, DBLP:conf/miccai/YangZCZC17, Chmelik201876}. In this paper, we only study probabilistic based data-driven active learning~\cite{Casanova2020Reinforced}. We train an initial model with a small set of labeled images, then new images for labeling are selected by the model using acquisition functions. The acquisition functions rely on uncertainty estimates from the model. Our model assumes that the highly uncertain images are more informative for improving segmentation accuracy. This process is repeated until the model reaches a certain accuracy. 

\begin{figure}[t!]
    \centering
    \includegraphics[width = 0.47\textwidth]{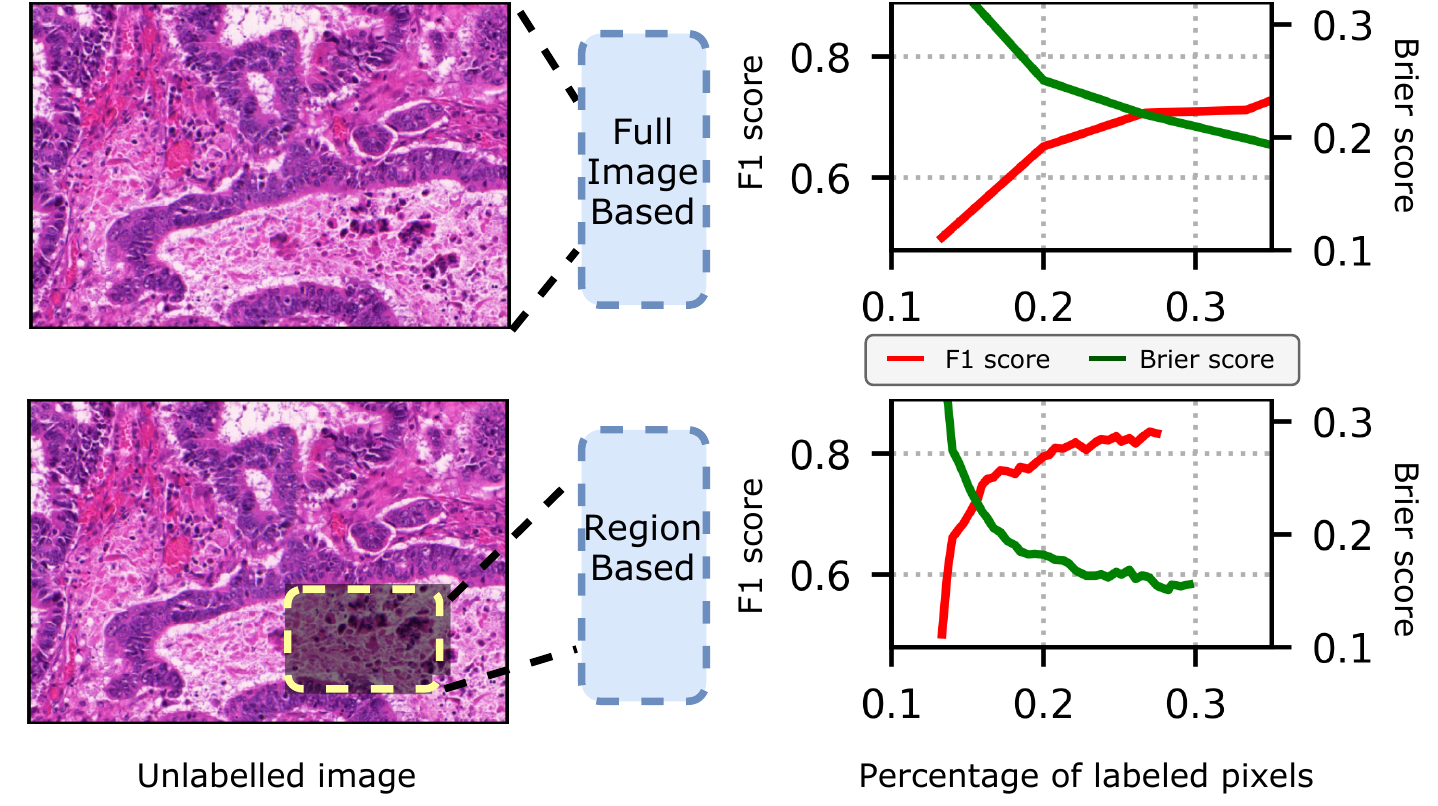}
    \caption{The comparison between full image based and region based acquisition strategies on GlaS dataset. Region acquisition can lead to higher segmentation accuracy (F1 score) and better calibrated model (Brier score) much faster than full image acquisition}
    \label{fig:first_figure}
\end{figure}

As acquisition functions, we use three well-known methods: VarRatio, Entropy and BALD~\cite{DBLP:conf/icml/GalIG17, Beluch_2018_CVPR}. We use two different acquisition strategies: full images annotation and region-based annotation. Annotating full images is the current trend where the oracle needs to label the entire image~\cite{DBLP:conf/miccai/YangZCZC17}. As for the region based approach, we here acquire square-shaped regions to be labeled~\cite{DBLP:conf/bmvc/MackowiakLGDLR18}. Region annotation has been shown to boosts the model performance compared to full image annotation~\cite{DBLP:journals/corr/abs-1911-11789, DBLP:conf/bmvc/MackowiakLGDLR18}, and in our particular study we demonstrate a reduction of annotation of pixels of up to 50\%, see Figure~\ref{fig:first_figure}.

Besides, we also shed some light on why region based active learning has better performance. We find that the uncertainty estimates from the models are not well-calibrated when it is trained with only a small amount of labeled data. The region based annotation can be more selective about pixels chosen for annotation and this leads to more well-calibrated models faster, see Figure~\ref{fig:first_figure}.
\subsection{Related work}
\textbf{Uncertainty used for AL} Different techniques have been investigated to represent the sample informativeness using the estimated uncertainty. A common approach is to combine an ensemble of several independent networks to estimate the uncertainty~\cite{Beluch_2018_CVPR, DBLP:conf/nips/SnoekOFLNSDRN19}. However, training multiple networks is computationally expensive. As an alternative, dropout has been proposed to obtain the posterior uncertainty over the network predictions~\cite{DBLP:conf/icml/GalIG17}.

\textbf{Acquisition strategies for AL}
Active learning for image segmentation is one of the lesser-explored computer vision tasks. The requirement of pixel-wise annotation allows for different acquisition strategies. Full image acquisition is the current trend~\cite{DBLP:conf/miccai/YangZCZC17}. Recently, \cite{DBLP:journals/corr/abs-1911-11789} proposed to use super-pixel segmentation to guide the model for selection and this demonstrated better performance. However, this highly depends on the quality of the super-pixel segmentation. An alternative to superpixels is to use fixed regions, and Mackowiak \emph{et al.} recently suggested a cost-effective region based active learning acquisition scheme that also outperformed full image annotation~\cite{DBLP:conf/bmvc/MackowiakLGDLR18}.

\textbf{Uncertainty calibration in AL} Even though a large number of studies have been carried out for improving uncertainty calibration on modern neural networks on tasks such as image classification~\cite{DBLP:conf/icml/GuoPSW17, DBLP:conf/nips/ThulasidasanCBB19} and anomaly detection~\cite{DBLP:conf/nips/SnoekOFLNSDRN19}, the relationship between uncertainty calibration and efficiency of active learning is rarely explored~\cite{Beluch_2018_CVPR}. Thus, in this paper, we apply multiple calibration quantification metrics to investigate the relationship between uncertainty calibration and performance of active learning for image segmentation tasks.  

\section{Method}
\label{sec:methodology}
\begin{figure*}
\centering
\resizebox{1.0\textwidth}{!}
{\begin{minipage}{\textwidth}
\includegraphics[width = 1.0\textwidth]{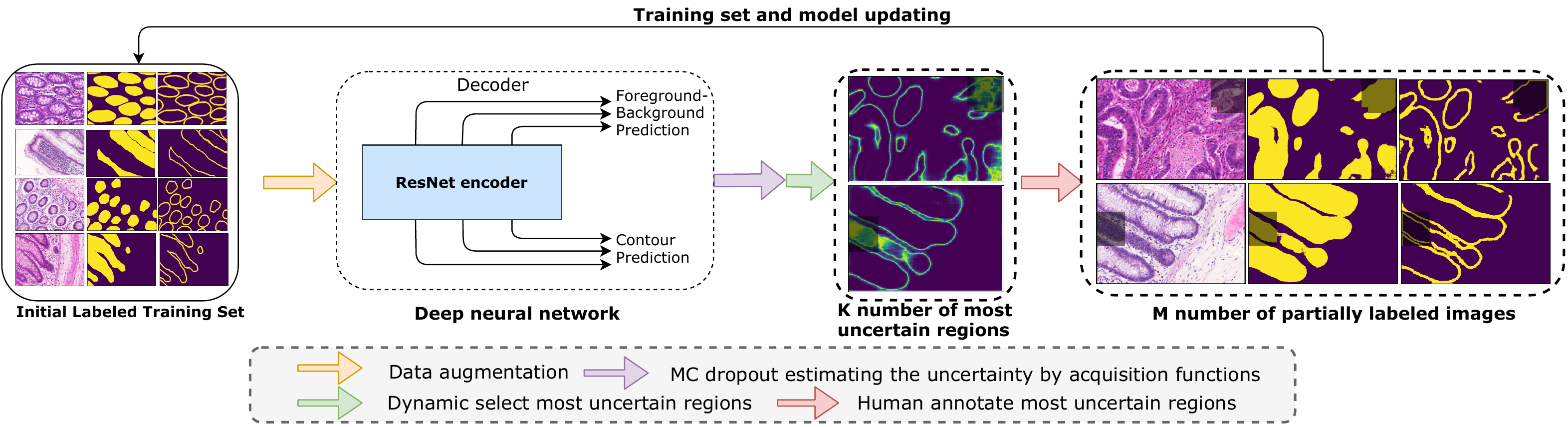}
\end{minipage}}
\caption{Region active learning framework. We train a model with the initial labeled images for segmentation. Then this model estimates the uncertainty of fixed-size regions in the unlabeled pool set and selects $K$ regions for the oracle to annotate. The model predictions for remaining pixels in the selected images are used as pseudo-labels to guide the oracle for labeling. After that, these images are sent back to the training set and used to update the model. This process is repeated until the model reaches a certain accuracy on the test data.}
\label{fig:framework}
\end{figure*}
This section consists of three major components: 1) the architecture of neural network 2) uncertainty estimation methods and acquisition strategies 3) quantification of uncertainty calibration.

\subsection{Architecture}
\label{sec:architecture}
Inspired by recent advances in multi-task learning~\cite{DBLP:conf/cvpr/0011QYH16} and residual neural networks~\cite{DBLP:conf/cvpr/HeZRS16, DBLP:conf/eccv/HeZRS16}, we utilize an architecture that uses ResNet-50~\cite{DBLP:conf/eccv/HeZRS16} as the feature extractor and employs the decoder of the DCAN model~\cite{DBLP:conf/cvpr/0011QYH16} to accomplish the image segmentation with auxiliary tasks~\cite{DBLP:journals/corr/WangLTL15a,DBLP:conf/aistats/LeeXGZT15,DBLP:conf/iccv/XieT15}. To estimate the uncertainty over the network predictions, dropout layers are added before each residual block in the feature extractor. We use the same loss function as~\cite{DBLP:conf/cvpr/0011QYH16} to train the model. As for quantifying segmentation accuracy, we use the commonly used evaluation metrics F1 score, object-level Dice index~\cite{DBLP:conf/cvpr/0011QYH16} and Jaccard index~\cite{DBLP:conf/isbi/CodellaGCHMDKLM18}. See Appendix section~\ref{sec:implementation_detail} for a detailed description of the architecture.

\subsection{Uncertainty estimation}
\label{sec:acquisition_func}
One approach for choosing informative areas for annotation is to evaluate the uncertainty of the estimation. We use Monte-Carlo dropout (MC-dropout) to approach the problem by interpreting dropout regularization as a variational Bayesian approximation~\cite{DBLP:conf/icml/GalG16}. In practice, we train the neural network with training data $\mathcal{D}_{train}$ using dropout, and prediction is done by performing $T$ stochastic forward passes through the network. In each stochastic forward pass $t$, a new dropout mask is generated which results in the weight $w_t$. These $T$ softmax vectors are then averaged thus obtaining the final posterior probability for pixel $x$ given class $c$:
\begin{equation}
    P(y=c|x) = \frac{1}{T}\sum_{t=1}^{T}P_t(y=c|x,w_t)
\end{equation}
We apply three different acquisition functions that have previously been demonstrated successful for active learning in 
the image classification problem~\cite{DBLP:conf/icml/GalIG17}: \textbf{VarRatio}, \textbf{Entropy}, and \textbf{BALD}. They are all based on the estimated uncertainty from the MC-dropout. A detailed explanation of these methods is in Appendix section~\ref{sec:acq_func}. Random selection is used as a baseline comparison.

\subsection{Acquisition strategies}
\label{sec:acquisition_strategy}
We use two different acquisition strategies: full image acquisition and region acquisition. The full image acquisition follows a standard active learning loop as described in most related literature~\cite{DBLP:conf/icml/GalIG17, Beluch_2018_CVPR} that the model requires oracles to annotated all the pixels in the image. The framework for the region acquisition can be seen in Figure~\ref{fig:framework}. The model learns to acquire annotation of pixels in the most uncertain regions instead of the full image. Besides, the prediction of the remaining pixels can be used to guide the oracle for labeling the pixels in the required regions. However, only the loss for the human-annotated regions is back-propagated. A detailed description for both acquisition loops is in Appendix section~\ref{sec:full_im_algo}. 

\subsection{Uncertainty calibration}
The described active learning framework relies on good uncertainty estimation since it represents the informativeness of images and regions with the estimated uncertainty. We thus inspect the uncertainty estimates of the aforementioned methods in each acquisition step using a variety of metrics; negative loglikelihood (NLL), expected calibration error (ECE), and the Brier score together with its decomposed version~\cite{DBLP:conf/nips/SnoekOFLNSDRN19}. See the Appendix section~\ref{sec:uncert_quantification} for a detailed explanation about these used metrics.

\section{Experiments}
\subsection{Datasets}
We use two datasets to evaluate the performance of the described framework:
\begin{itemize}
    \item GlaS Challenge Contest~\cite{DBLP:journals/mia/Sirinukunwattana17} contains 85 training images (37 benign and 48 malignant) and 80 test images with ground truth annotations provided by an pathologist. All images are with size $528$x$784$. 
    \item 2016 International Skin Imaging Collaboration (ISIC)~\cite{DBLP:conf/isbi/CodellaGCHMDKLM18} contains 900 training images and 379 test images. The image size ranges from $542$x$718$ to $2048$x$2048$. All the images are bi-linearly down-sampled to $384$x$512$. This specific ratio is chosen since most of the images a height to width ratio of $3$:$4$. 
\end{itemize}
Neither datasets has contours so these are extracted with a Sobel filter and dilated with a disk filter (radius = 3). 

\subsection{Results}
\begin{figure}[t]
    \subfigure[]{\includegraphics[width=0.235\textwidth]{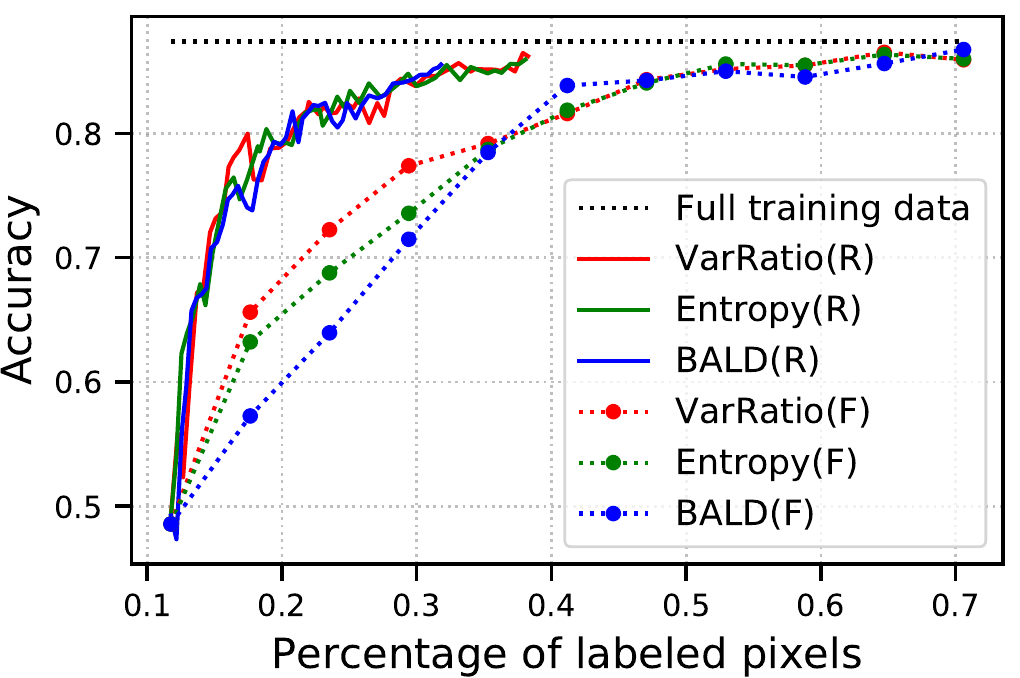}}\hspace{0.005cm}
    \subfigure[]{\includegraphics[width=0.235\textwidth]{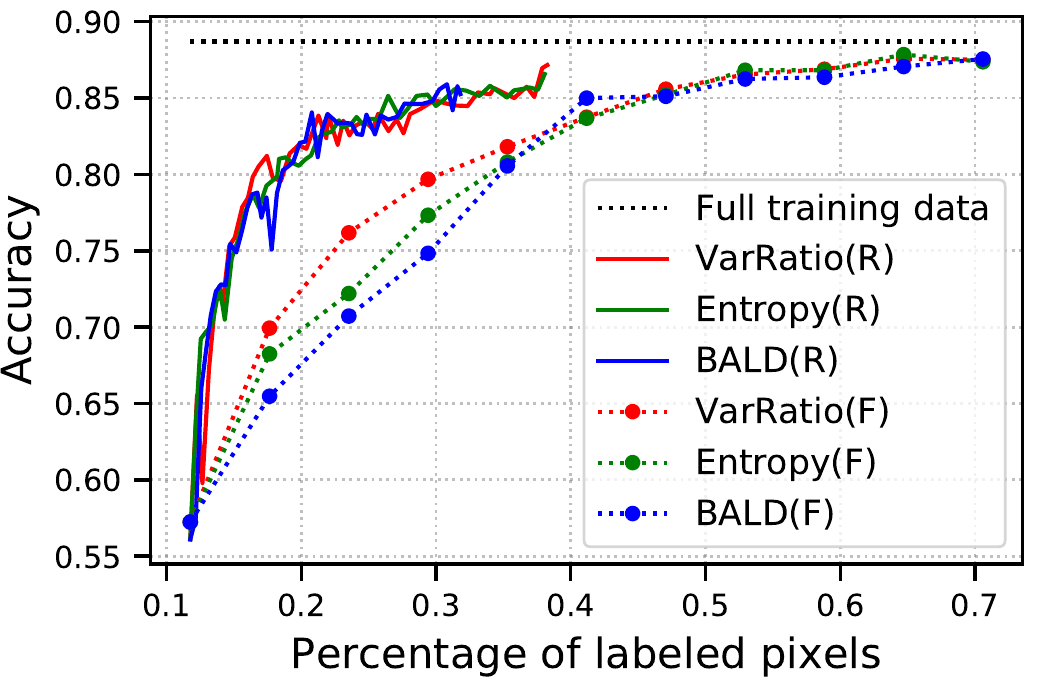}}\hspace{0.005cm}
    \subfigure[]{\includegraphics[width=0.235\textwidth]{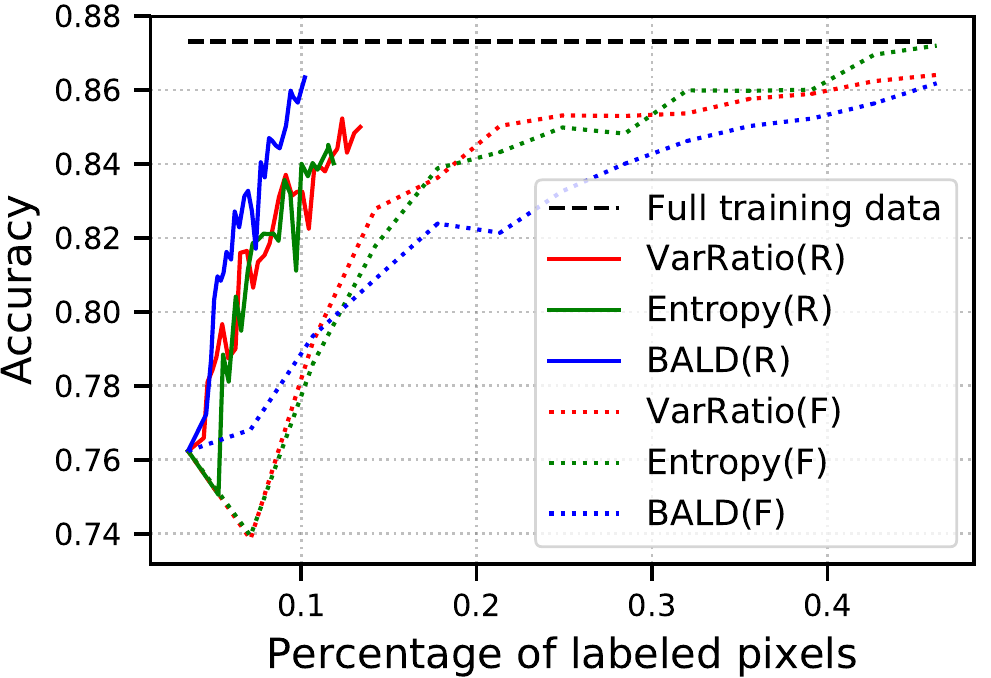}} 
    \subfigure[]{\includegraphics[width=0.235\textwidth]{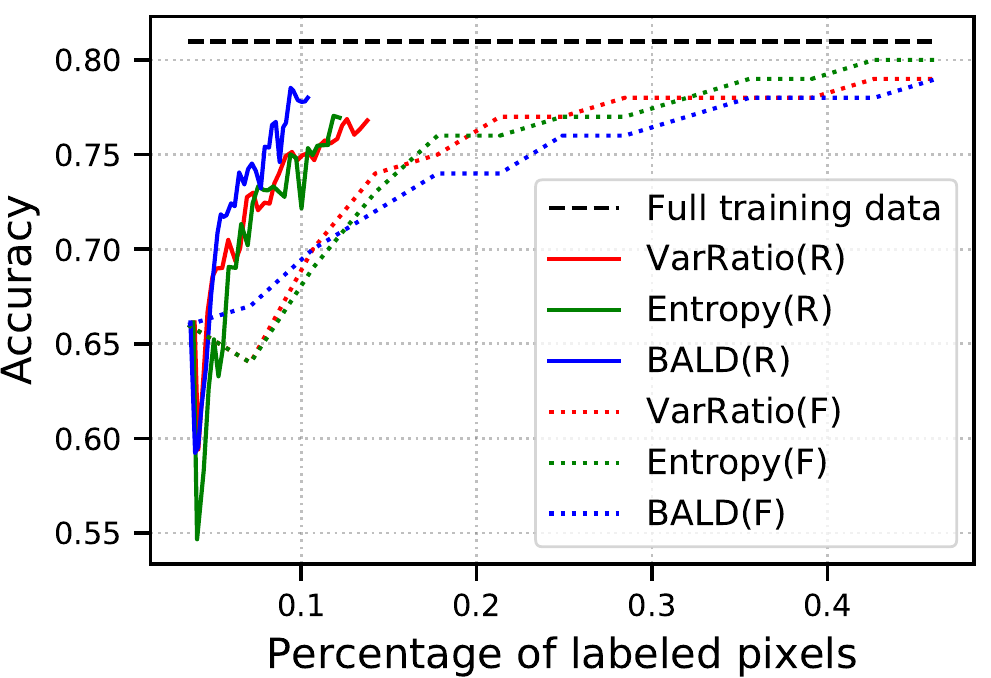}} 
    \caption{Segmentation accuracy F1 score (a) and dice index (b) on GlaS dataset, and dice index (c) and Jaccard index (d) on ISIC dataset using full image (\textit{F}) and region (\textit{R}) acquisition strategies. \textit{Full training data} represents the performance of the model when it is trained with all the training images. Compared to the acquisition of full images, acquiring regions can significantly reduce the labeling cost while reaching a similar accuracy.}
    \label{fig:accuracy_plot}
\end{figure}

Figure~\ref{fig:accuracy_plot} shows the increment of averaged segmentation accuracy over four runs on both datasets as more images added to the training set. The region acquisition strategy decreases the amount of required annotated pixels up to a factor of two compared to the full image acquisition strategy for reaching the limiting performance. For the region-based acquisition, the BALD acquisition function further cuts the labeling effort on the ISIC dataset compared to another two methods. However, the effect of annotation strategy far out-weights the effect acquisition function choice. 
\begin{figure}[ht]
    \centering
    \includegraphics[width = 0.47\textwidth]{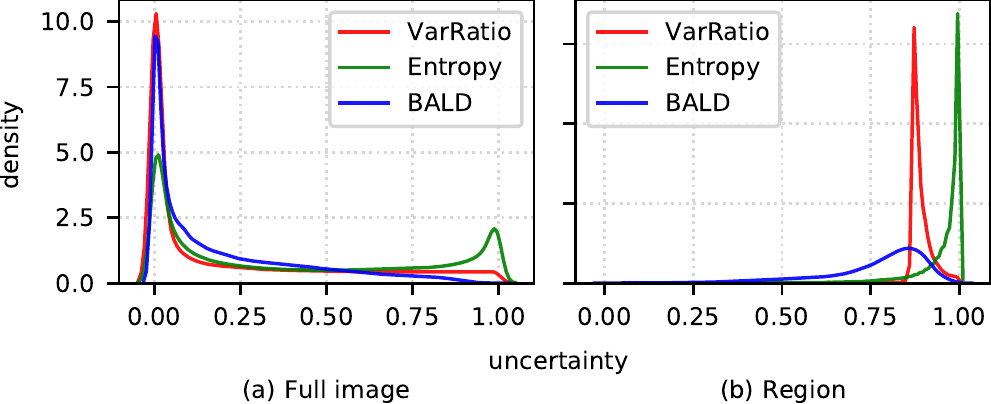}
    \caption{Normalized Uncertainty distribution of pixels in the acquired images using full image strategy (a) and regions using region acquisition strategy (b) at the first acquisition step. The majority pixels in the selected images have low uncertainty, whereas the region acquisition strategy can highly reduce the ratio of this type of pixels in the selection.}
    \label{fig:uncert_distribution}
\end{figure}

To shed some light on why region annotation is more efficient than the full image annotation, we explore how uncertainty estimates are distributed in the images selected for annotation.
The observed and expected segmentation accuracy was calculated to assess the calibration quality. Plotting both values against each other revealed that the model tends to be overconfident at the beginning of the acquisition process (see Figure~\ref{fig:ece_histogram} in the Appendix). The uncertainty distribution in the selected images and regions are shown in Figure~\ref{fig:uncert_distribution}. The majority of the pixels in the selected images have fairly low uncertainty which means that the model is querying the oracle to label a high amount of pixels that have high chances of already being correctly predicted by the model. The region acquisition strategy mitigates this by selecting regions for annotation that have a smaller ratio of certain pixels, see Figure~\ref{fig:uncert_distribution} (b).

\begin{figure}[t!]
    \centering
    \includegraphics[width=.41\textwidth]{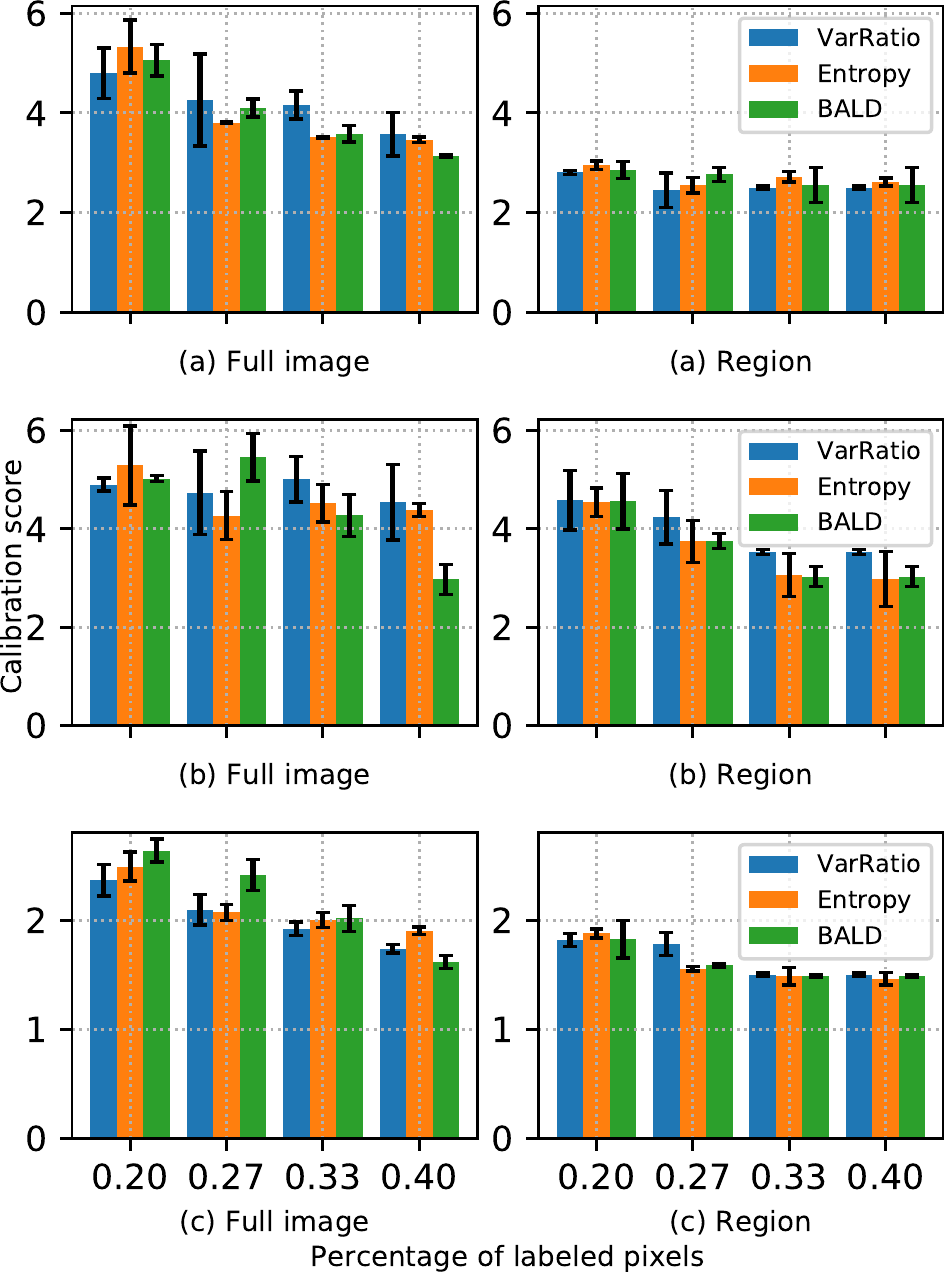}
    \caption{Uncertainty calibration at different acquisition steps under multiple acquisition functions and strategies on the GlaS dataset. The used evaluation metrics are (a) NLL, (b) ECE, and (c) Brier score. Region acquisition can much faster lead to a better calibrated model than acquiring full images no matter which uncertainty estimation method is used.}
    \label{fig:overall_calibration}
\end{figure}

Figure~\ref{fig:overall_calibration} displays the overall comparison of uncertainty calibration at several acquisition steps over all the evaluation metrics. See the Appendix for the uncertainty calibration at each acquisition step. Region annotation much faster leads to a well-calibrated model compared to full image annotation regardless of the evaluation metric. Specifically, the BALD and Entropy acquisition functions lead to a better ECE score compared to VarRatio. The change of the uncertainty calibration quality in Figure~\ref{fig:overall_calibration_all_acq_step} follows the increment of the segmentation accuracy as shown in Figure~\ref{fig:accuracy_plot}. It indicates that a well-calibrate model can further boost the efficiency and effectiveness of the active learning acquisition process. 

\section{Conclusion}
We carried out a study on using different uncertainty estimation methods and acquisition strategies in active learning for the image segmentation task. We empirically showed that annotating regions can significantly reduce the labeling effort and boost the effectiveness of the active learning framework compared to full image annotating. Besides, we observe that region acquisition strategies much faster leads to a better calibrated model than full image acquisition strategies no matter which uncertainty estimation method is used. This provides an explanation for the superiority of the region acquisition active learning.
\section*{Acknowledgements}
This research was funded by the IDUN Center of Excellence supported
by the Danish National Research Foundation (DNRF122) and the Velux
Foundations (Grant No. 9301). We thank NVIDIA Corporation for donating GPU. We thank Søren Hauberg for the fruitful discussions.

Code and data files can be found at~\citep{DTUdata}.
\newpage
\bibliography{icmlbib}
\bibliographystyle{icml2020}

\clearpage
\newpage
\appendix
\counterwithin{figure}{section}
\section{A review of uncertainty estimation methods}
\label{sec:acq_func}
We apply three different MC dropout oriented uncertainty estimation methods: \textbf{VarRatio}, \textbf{Entropy} and \textbf{BALD}. The VarRatio acquisition function choose points that the model is 
\textit{least certain} for the predictions (can also be called as \textbf{Variation Ratios}):
\begin{equation}
    \text{VarRatio}[x] = 1-\max_{y}P(y|x)
\end{equation}
A more generalized method is to evaluate the \textbf{Entropy} of the prediction. It is not restricted to the 
binary case and can be used directly in the multi-class scenario as well:
\begin{equation}
    \text{Entropy}[x] = -\sum_{j}P(y=j|x)\log P(y=j|x)
\end{equation}
The acquisition functions above only encodes the relative uncertainty between different classes. However, \textbf{BALD} 
considers both relative uncertainty and model uncertainty. \textbf{BALD} calculates the mutual information between the 
predictions and model posterior \cite{Houlsby2011BayesianAL, DBLP:conf/icml/GalIG17}:
\begin{equation}
\begin{aligned}
\text{BALD}[x] = & \text{Entropy}[x]+ \\ 
                & \sum_{t=1}^{T}\sum_{j}P_t(y=j|x,w_t)\log P_t(y=j|x,w_t)    
\end{aligned}
\end{equation}
Random selection (\textbf{Rand}) is performed the baseline comparison.

\section{Active learning loop}
\label{sec:full_im_algo}
\subsection{Full image annotation suggestion}
\label{subsec:full_im_algo}
The scenario where the oracle needs to label all the pixels in the suggested images by using the proposed active learning algorithm as shown in Alg.\ref{alg:sec_3_full_im_annotation}.

\begin{algorithm}
    \caption{Active learning with full image annotation suggestion}
    \label{alg:sec_3_full_im_annotation}
    \begin{algorithmic}[1]
    \REQUIRE Labeled Set $\mathcal{L}$, Unlabeled Pool Set $\mathcal{U}$
    \STATE Train initial CNN model with the same training data $\mathcal{L}$ multiple times, and select the best one 
    based on the validation loss as starting point
    \FOR {Number of query steps to make from pool set $\mathcal{U}$}
        \STATE Estimate the uncertain value for per pixel $x$ in all images $\mathcal{X}$ in the unlabeled pool set 
        $\mathcal{U}$ using acquisition functions from Sec. \ref{sec:acq_func}
        \STATE Calculate the utility score for per image $\mathcal{X}$ by adding the uncertain value for all the 
pixels 
        $x$ from that image and select $M$ most informative images $\mathcal{X}^*$.
        \STATE Annotating all the pixels in the suggested images $\mathcal{X}^*$, and add them with the corresponding 
        label $Y^*$ to $\mathcal{L}$. Then remove $\mathcal{X}^*$ from pool set $\mathcal{U}$
        \STATE Retrain the CNN model with the updated training set $\mathcal{L}$ until it converge
        \STATE Evaluate the new trained CNN model on test dataset
    \ENDFOR
    \end{algorithmic}
\end{algorithm}
We train the initial network model multiple times with the small labeled set $\mathcal{L}$ and select the best one based on the validation loss as the starting point for the query process. This is done to isolate the effects of the acquisition functions and to avoid solutions in bad local minima. In each acquisition step, the model selects the most uncertain unlabeled images from the pool and queries the oracle to annotate all the pixels in the selected images. The images are then moved from the pool set $\mathcal{U}$ to the labeled set $\mathcal{L}$. Following that, a new CNN model is trained from scratch with the new training set $\mathcal{L}$. This process is repeated until model performance has converged.
\subsection{Region specific annotation suggestion}

We use the same approach as Sec.\ref{subsec:full_im_algo} to estimate the uncertainty $\mathcal{Q}(x)$ for all the pixels $x$ in each unlabeled image $\mathcal{X}$. Then the images, which include the $M$ most uncertain regions, with their corresponding pseudo-labels are added to the training data $\mathcal{L}$ as shown in Alg.\ref{alg:sec_3_selec_most_uncert_region}. These already selected images are still kept in the pool set since it is possible for the model to suggest another region in previously selected images for the annotation. In addition, it is assumed that the annotations are error-free so the uncertainty value for pixels in $\mathcal{R}^*$ are manually assigned as zero. Following that, the CNN model is re-trained from scratch with the updated annotated data $\mathcal{L}$ four times and the best model is selected based on validation performance for the next acquisition step. This is done due to avoid bad local minima. It is also worth to mention that only the loss for the human annotated pixels are back-propagated during the training process. 

\begin{algorithm}
    \caption{Search and select the most uncertain regions}
    \label{alg:sec_3_selec_most_uncert_region}
    \begin{algorithmic}[1]
    \REQUIRE
        \STATE Uncertainty estimation $\mathcal{Q}(x)$ for all the unlabeled images $\mathcal{X}$ in the unlabeled 
        pool set $\mathcal{U}$
        \STATE Kernel window with size $[k_w, k_h]$ and value $k_v$
    \IF{There exists already selected regions $\mathcal{R}^*$}
    \STATE $\mathcal{Q}[x \in \mathcal{R}^*]$ = 0
    \ENDIF
    \FOR {Each image from pool set $\mathcal{U}$}
        \STATE Convolve the uncertainty estimation map $\mathcal{Q}(x)$ with stride size $k_s$ and kernel window to 
        obtain the uncertainty score for each region
    \ENDFOR
    \STATE Rank all the regions based on their uncertainty score and select the $M$ most uncertain regions 
    $\mathcal{R}^*$. The corresponding images are denoted as $\mathcal{X}^*$
    \STATE Require oracle to annotate the pixels in the selected regions $\mathcal{R}^*$. The remaining pixels in the 
    selected images $\mathcal{X}^*$ are assigned pseudo-labels by using the predictions from the model. 
    \STATE Add the selected images $\mathcal{X}^*$ and pseudo-labels into labeled set $\mathcal{L}$
    \end{algorithmic}
\end{algorithm}

\section{Uncertainty calibration}
\label{sec:uncert_quantification}
We use the commonly used negative log likehood (NLL), expected calibration error (ECE) and Brier score to quantify the uncertainty calibration.

\textbf{Negative log likelihood} is a standard approach to measure probabilistic models' quality. It can be formulated as:
\begin{equation}
    \mathcal{L} = - \sum_{i=1}^{n}\log (\hat{\pi}(y_i|x_i))
\end{equation}
where $\hat{\pi}(Y|X)$ is the probabilistic model. Although this is a commonly used metric, it tends to over-emphasize the tail probabilities~\cite{DBLP:conf/mlcw/CandelaRSBS05}.

\textbf{Expected calibration error} summarizes the information in the reliability diagram and measures the difference in expectation between confidence and accuracy as follows: we first group the predictions into $K$ bins with equal size, let $B_k$ be the set of the samples whose predicted probability $\hat{p}_i$ (the maximum value from the softmax output) fall into bin $k$, then the accuracy and confidence for $B_k$ is:
\begin{subequations}
\begin{equation}
    \text{acc}(B_k) = \frac{1}{|B_k|}\sum_{i\in B_k} 1(\hat{y}_i = y_i)
\end{equation}
\begin{equation}
    \text{conf}(B_k) = \frac{1}{|B_k|}\sum_{i\in B_k} \hat{p}_i
\end{equation}
\end{subequations}
The ECE score is calculated as:
\begin{equation}
    \text{ECE} = \sum_{k=1}^{K}\frac{|B_k|}{n} |\text{acc}(B_k) - \text{conf}(B_k)|
\end{equation}

\textbf{Brier score} is another metric that measures the quality of the predicted probability. It measures the difference between the predicted probability $\hat{p}_i$ and the one-hot encoded ground truth $y_i$:
\begin{equation}
    \text{BS} = |\mathcal{Y}^{-1}|(1 - 2p_{i}+\sum_{y\in \mathcal{Y}}p_i^2)
\end{equation}
The brier score is appropriate for binary and categorical outcomes, and the lower Brier score for a set of predictions, the better the predictions are calibrated. According to~\cite{DBLP:conf/csoc/QasemFRS17}, the brier score can also be decomposed into three components: \textit{reliability} which measures how close the predicted probability are to the true probability, \textit{uncertainty} which measure the inherent uncertainty in the dataset and \textit{resolution} represents the deviation of individual predictions against the marginal. Therefore, we also use the reliability from the decomposed brier score to quantify the uncertainty calibration.

\begin{table}[t!]
\resizebox{0.85\textwidth}{!}
{\begin{minipage}{\textwidth}
\begin{tabular}{@{}lllll@{}}
\toprule
 Accuracy & Rand & VarRatio & Entropy & BALD \\ 
 & \multicolumn{4}{l}{Active full image suggestion (Region suggestion)} \\ \midrule \midrule
 \multicolumn{5}{c}{\begin{tabular}[c]{@{}l@{}}F1 score on GlaS dataset\end{tabular}} \\ \midrule
\begin{tabular}[c]{@{}l@{}}75\% mean\\ std\end{tabular} & \begin{tabular}[c]{@{}l@{}}29.41 (-)\\ 2.17 (-)\end{tabular} & \begin{tabular}[c]{@{}l@{}}29.41 (\textbf{16.04})\\ 2.34 (\textbf{0.78})\end{tabular} & \begin{tabular}[c]{@{}l@{}}35.29 (\textbf{15.83})\\ 3.40 (\textbf{1.61})\end{tabular} & \begin{tabular}[c]{@{}l@{}}37.5 (\textbf{16.33})\\ 2.59 (\textbf{2.55})\end{tabular} \\
\begin{tabular}[c]{@{}l@{}}80\% mean\\ std\end{tabular} & \begin{tabular}[c]{@{}l@{}}41.18 (-)\\ 2.31 (-)\end{tabular} & \begin{tabular}[c]{@{}l@{}}35.29 (\textbf{17.45})\\ 2.03 (\textbf{0.21})\end{tabular} & \begin{tabular}[c]{@{}l@{}}41.18 (\textbf{18.84})\\ 2.58 (\textbf{1.26})\end{tabular} & \begin{tabular}[c]{@{}l@{}}41.18 (\textbf{20.3})\\ 1.18 (2.55)\end{tabular} \\
\begin{tabular}[c]{@{}l@{}}85\% mean\\ std\end{tabular} & \begin{tabular}[c]{@{}l@{}}76.47 (-)\\ 1.11 (-)\end{tabular} & \begin{tabular}[c]{@{}l@{}}52.94 (\textbf{28.82})\\ 1.04 (\textbf{0.15})\end{tabular} & \begin{tabular}[c]{@{}l@{}}52.94 (\textbf{29.39})\\ 1.12 (\textbf{0.89})\end{tabular} & \begin{tabular}[c]{@{}l@{}}52.94 (\textbf{31.23})\\ 1.55 (\textbf{0.44})\end{tabular} \\
\midrule
\midrule
\multicolumn{5}{c}{\begin{tabular}[c]{@{}l@{}}Jaccard index on ISIC dataset\end{tabular}} \\ \midrule
\begin{tabular}[c]{@{}l@{}}70\% mean\\ std\end{tabular} & \begin{tabular}[c]{@{}l@{}}10.67 (-)\\ 1.75 (-)\end{tabular} & \begin{tabular}[c]{@{}l@{}}10.67 (\textbf{5.34})\\ 2.51 (\textbf{0.31})\end{tabular} & \begin{tabular}[c]{@{}l@{}}14.22 (\textbf{6.58})\\ 2.17 (\textbf{1.84})\end{tabular} & \begin{tabular}[c]{@{}l@{}}14.22 (\textbf{4.93})\\ 2.64 (\textbf{0.35})\end{tabular} \\
\begin{tabular}[c]{@{}l@{}}77\% mean\\ std\end{tabular} & \begin{tabular}[c]{@{}l@{}}46.22 (-)\\ 0.18 (-)\end{tabular} & \begin{tabular}[c]{@{}l@{}}21.33 (\textbf{13.78})\\ 1.65 (\textbf{1.16})\end{tabular} & \begin{tabular}[c]{@{}l@{}}28.44 (\textbf{13.12})\\ 1.96 (\textbf{1.39})\end{tabular} & \begin{tabular}[c]{@{}l@{}}32.00 (\textbf{9.55})\\ 1.10 (\textbf{0.39})\end{tabular} \\
\bottomrule
\end{tabular}
\end{minipage}}
\\
\caption{To achieve a certain segmentation accuracy, the mean and standard deviation of the amount of required pixels that need to be annotated using different uncertainty estimation methods and acquisition strategies. Region acquisition strategy can significantly reduce the labeling effort compared to full image acquisition strategy.}
\label{tab:sec_region_specific_anno_mean_std}
\end{table}

\section{Implementation details}
\label{sec:implementation_detail}
The network architecture is identical regardless of the datasets. We use three different encoder structures: resnet-v2-50, resnet-v2-101 and resnet-v2-152. We train the model with the loss function as shown in Eq.~\ref{eq_loss}, where the first term is a regularization term, and $\mathcal{L}_{auxiliary}$ and $\mathcal{L}_e$ represent the segmentation loss from the auxiliary predictions and final predictions. The detailed architecture and performance for this segmentation model are demonstrated in the thesis~\cite{li2018a}. To check the performance of the proposed active learning frameworks, only resnet-v2-50 is used since it has the lowest number of parameters and had comparable performance to resnet-v2-101 and resnet-v2-152. 
\begin{equation}
    \mathcal{L}_{total} = \lambda \psi(\theta) + w\mathcal{L}_{auxiliary} + \mathcal{L}_{e}
    \label{eq_loss}
\end{equation}

As for the settings for the active learning loop, we randomly select 10 images from the training set as the initial training data. For the full image annotation framework, we select 5 images in each acquisition step. All experiments are run until the performance of the final model is converged regardless of the extra annotation effort on unlabeled pool data. For region-based annotation, we acquire regions until the model achieve a certain accuracy. 

\section{Performance}
The segmentation accuracy at each acquisition step using different uncertainty estimation methods and random selection are shown in Table~\ref{tab:sec_region_specific_anno_mean_std}. As for the full image acquisition strategies, the uncertainty estimation based acquisition functions perform worse than random selection in the beginning. However, they tend to significantly outperform random selection as more images are added into the training set. Besides, the region acquisition strategy can highly reduce the labeling effort compared to the full image acquisition strategy no matter which uncertainty estimation method or dataset is used. Furthermore, acquiring regions with BALD can further cut the labeling effort compared to other two uncertainty estimation methods on ISIC dataset.

\begin{figure}[t]
    \centering
    \includegraphics[width=.4\textwidth]{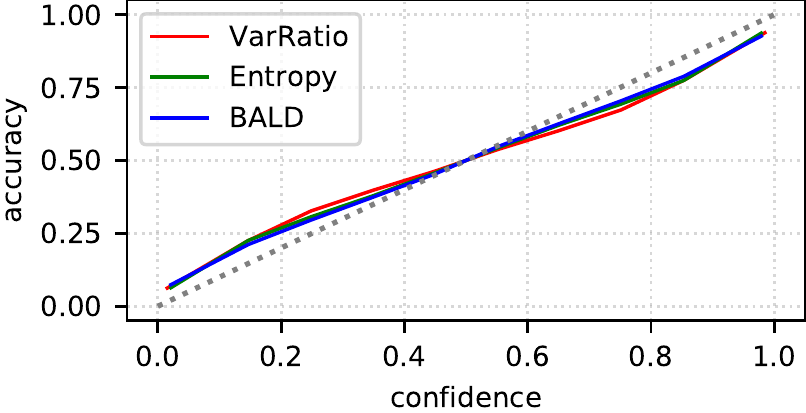}
    \caption{Reliability diagram for the first step using full image acquisition strategy. The model is not well-calibrated in the beginning of the acquisition process.}
    \label{fig:ece_histogram}
\end{figure}
\begin{figure}[!t]
    \centering
    \includegraphics[width=.40\textwidth]{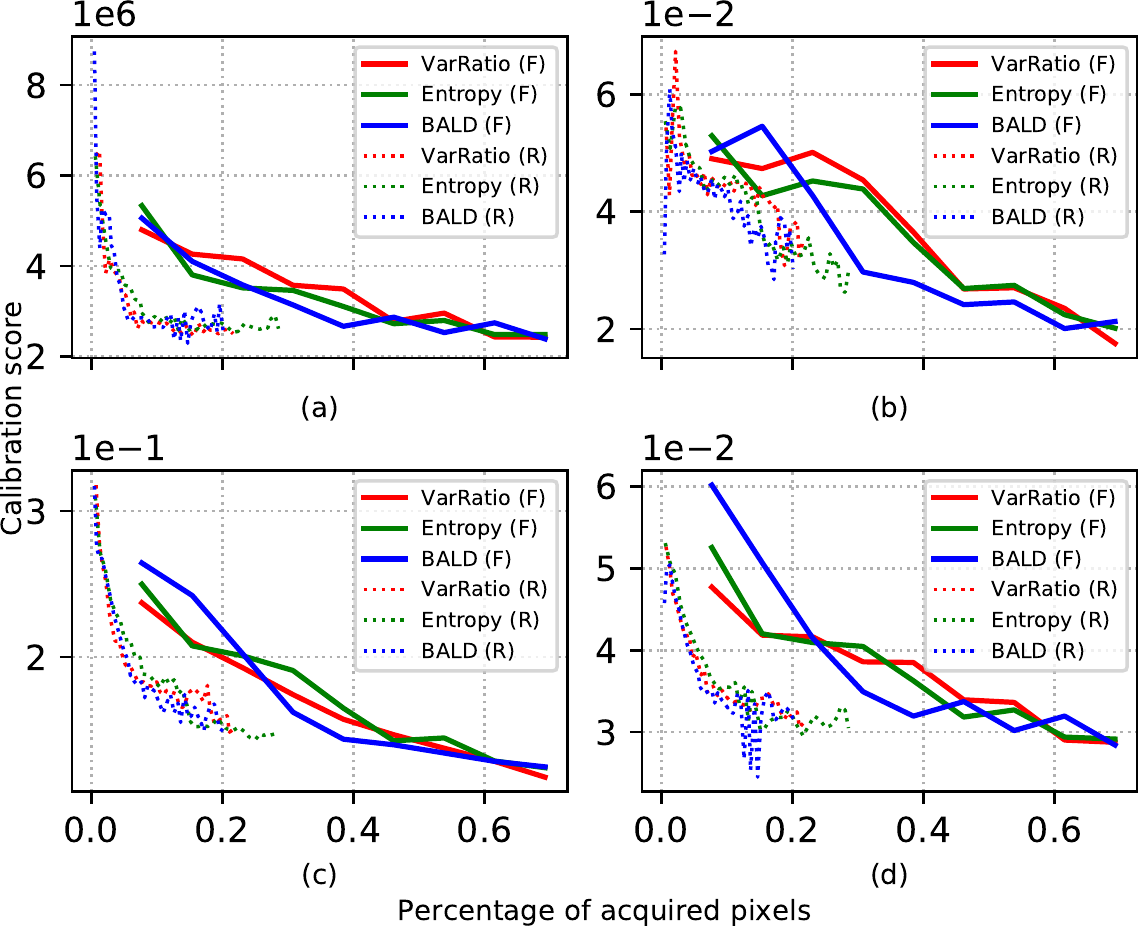}
    \caption{Uncertainty calibration under all types of uncertainty estimation methods and acquisition strategies on GlaS dataset. The used evaluation metrics from (a) to (d) are NLL, ECE, Brier score and reliability from the decomposed Brier score. Acquire regions can much faster lead to a better calibrated model than acquiring full images no matter which uncertainty estimation method is used}
    \label{fig:overall_calibration_all_acq_step}
\end{figure}

In addition, we also show the uncertainty calibration through the whole acquisition process. Figure~\ref{fig:ece_histogram} illustrates that the model is not well-calibrated in the beginning of the acquisition process and it tends to be overconfident. The detailed quantification of the uncertainty calibration is shown in Figure~\ref{fig:overall_calibration_all_acq_step}. The movement of the uncertainty calibration quality follows the increment of the segmentation accuracy which indicates that a better calibrated model can boost the efficiency of active learning further.

\end{document}